\documentclass[11pt]{article}
\usepackage{acl}
\usepackage{times}
\usepackage{latexsym}
\usepackage[T1]{fontenc}
\usepackage[utf8]{inputenc}
\usepackage{microtype}
\usepackage{graphicx}
\usepackage{subcaption}
\usepackage{booktabs}
\usepackage{amsfonts}
\usepackage{nicefrac}
\usepackage{amsmath}
\usepackage{amssymb}
\usepackage{enumitem}
\usepackage{makecell}
\usepackage{float}
\usepackage{multirow}
\usepackage{algorithm}
\usepackage{algorithmic}
\usepackage[capitalize,noabbrev]{cleveref}

\title{From Errors to Rules: Iterative Prompt Optimization for Text Classification}

\author{Yueying Cui \quad Renhao Xue \quad Yi Zhang \quad Mukul Prasad \\
  Amazon Web Services}

\begin{document}
\maketitle

\begin{abstract}
Prompt optimization for text classification spans diverse approaches, from demonstration selection to exploration-based search to error-driven diagnosis, each with known but incompletely characterized strengths and limitations. We conduct a comprehensive empirical study across diverse classification benchmarks (2 to 150 classes) comparing these paradigms through both quantitative evaluation and qualitative analysis of optimization traces, revealing that each paradigm excels on structurally different task types and that no single method dominates. Guided by these insights, we propose Error-Guided Optimization (ERGO), an error-driven method that iterates over the full training set in non-overlapping batches, diagnoses classification failures, and generates targeted decision rules through a diagnose-prescribe-rewrite feedback loop. ERGO achieves the best accuracy on tasks where errors concentrate in specific confused label pairs (which we term boundary-learnable tasks): TREC: 90.0\%, CLINC150: 94.4\%, converges in 3-5 iterations, and produces interpretable decision rules. While ERGO does not achieve the highest overall average, it fills a complementary role: demonstration-based ICL wins on coverage-dependent tasks, exploration-based search wins on many-class intent, and ERGO wins where decision boundaries are learnable from error patterns. We provide a complementarity framework linking task characteristics to optimal paradigm selection, offering practical guidance for practitioners.
\end{abstract}

%%%%%%%%%%%%%%%%%%%%%%%%%%%%%%%%%%%%%%%%%%%%%%%%%%%%%%%%%%%%%%%%%%%%%%%%%%%%%%%
\section{Introduction} \label{sec:intro}
%%%%%%%%%%%%%%%%%%%%%%%%%%%%%%%%%%%%%%%%%%%%%%%%%%%%%%%%%%%%%%%%%%%%%%%%%%%%%%%

Prompt optimization for text classification has produced a rich landscape of methods, from demonstration selection~\cite{liu2021makes} to exploration-based search~\cite{zhou2023largelanguagemodelshumanlevel,khattab2023dspycompilingdeclarativelanguage} to error-driven diagnosis~\cite{pryzant2023automatic}. Unlike fine-tuned models \cite{devlin2019bert,app14146254} that require substantial labeled data, these methods operate in the few-shot regime \cite{sahoo2025systematicsurveypromptengineering}, yet they embody fundamentally different optimization strategies. Existing comparisons typically focus on overall accuracy without analyzing when and why each approach excels \cite{li2025surveyautomaticpromptengineering,ramnath2025systematicsurveyautomaticprompt}. Recent findings that prompt optimization is ``statistically indistinguishable from a coin flip''~\cite{diagnosing2025} further underscore the need for a deeper understanding of which tasks benefit from which paradigm.

We address this gap through a comprehensive empirical study across diverse classification benchmarks (2--150 classes), comparing three paradigms: demonstration selection methods that choose diverse few-shot examples \cite{liu2021makes,yu2022generate}, exploration-based methods that search over instruction--demonstration combinations (DSPy \cite{khattab2023dspycompilingdeclarativelanguage}, GEPA \cite{agrawal2026gepa}, APE \cite{zhou2023largelanguagemodelshumanlevel}), and our proposed diagnosis-based method (ERGO) that uses classification errors to generate targeted decision rules. Our analysis reveals that these paradigms are complementary rather than competing: each excels on structurally different task types, and no single method dominates overall.

We propose Error-Guided Optimization (ERGO), which iterates over the full training set using a diagnose$\rightarrow$prescribe$\rightarrow$rewrite meta-prompt: at each iteration, ERGO classifies a batch, identifies confused label pairs, and generates explicit decision rules. Unlike exploration-based methods that treat accuracy as a black-box signal, ERGO jointly refines instructions, demonstrations, and classification guidelines from actual error patterns, producing interpretable rules that encode task-intrinsic knowledge.

ERGO provides unique value on boundary-learnable tasks, where classification errors concentrate in specific confused pairs amenable to natural-language rules. On such tasks, ERGO achieves the best accuracy (TREC: 90.0\%, CLINC150: 94.4\%), typically converges in 3-5 iterations, and produces rules that transfer across model families. For instance, a single misclassified query about a company triggers the rule ``organizations $\rightarrow$ human beings,'' fixing 25+ test queries. However, ERGO does not dominate overall: ICL-Diversity wins on coverage-dependent tasks at zero optimization cost, and DSPy wins on many-class tasks, confirming that the paradigms are complementary rather than competing.

Our main contributions:
\begin{itemize}[noitemsep, left=0pt]
\item ERGO, an error-driven prompt optimization method that produces interpretable, transferable decision rules through iterative feedback over the full training set (\cref{sec:method}).
\item A characterization of \emph{boundary learnability}, grounded in the observed error structure, that identifies where those rules pay off: whether errors express a recurring distinction (\emph{discovery}), whether it is pair-local, spread across many rare pairs, or diffuse (\emph{structure}), and whether acting on it helps without collateral damage (\emph{repairability}).
\item Case studies showing how ERGO discovers learnable decision boundaries that other methods miss: hidden conventions, progressive boundary refinement, and structural rules that generalize across label pairs (\cref{sec:case_studies_qual}).
\item Quantitative analysis showing ERGO generalizes across 5 models from 4 families and transfers prompts from stronger to weaker models without statistically significant accuracy loss (\cref{sec:case_studies}).
\item A practical paradigm selection framework, validated on 6 held-out datasets: start with ICL-Diversity, apply ERGO for boundary-learnable tasks, use DSPy for many-class tasks (\cref{sec:complementarity,sec:appendix-held-out}).
\end{itemize}

%%%%%%%%%%%%%%%%%%%%%%%%%%%%%%%%%%%%%%%%%%%%%%%%%%%%%%%%%%%%%%%%%%%%%%%%%%%%%%%
\section{Prompt Optimization Paradigms} \label{sec:landscape}
%%%%%%%%%%%%%%%%%%%%%%%%%%%%%%%%%%%%%%%%%%%%%%%%%%%%%%%%%%%%%%%%%%%%%%%%%%%%%%%

We organize prompt-based classification methods into three paradigms: selecting demonstrations without iterative optimization (Demonstrate), searching over prompt variants guided by accuracy signals (Explore), and systematically diagnosing classification errors to refine prompts (Diagnose).

\paragraph{Paradigm 1: Demonstrate.} Select diverse few-shot examples as demonstrations \cite{brown2020languagemodelsfewshotlearners,dong2022survey}. Research has explored selection strategies \cite{liu2021makes,margatina2023active}, ordering effects \cite{guo-etal-2024-makes}, and diversity-based retrieval \cite{yu2022generate}. In our setup, ICL-Diversity selects $k{=}20$ demonstrations via $k$-means on Sentence-BERT \cite{reimers2019sentence} embeddings \cite{su2023selective}; ICL-Uniform samples randomly \cite{brown2020languagemodelsfewshotlearners}. Neither involves iterative refinement or adapts to classification errors.

\paragraph{Paradigm 2: Explore.} Methods in this paradigm search over prompt variants guided by accuracy signals. APE \cite{zhou2023largelanguagemodelshumanlevel,honovich2022instructioninductionexamplesnatural} generates instruction candidates via LLM sampling. OPRO \cite{yang2024largelanguagemodelsoptimizers} uses LLM-driven meta-optimization. DSPy/MIPROv2 \cite{khattab2023dspycompilingdeclarativelanguage} explores instruction$\times$demonstration combinations via Bayesian optimization, selecting candidates based on 35-item minibatch scores. GEPA \cite{agrawal2026gepa} uses evolutionary search with per-example error feedback to guide instruction evolution. Other methods include PromptWizard \cite{agarwal2024promptwizardtaskawarepromptoptimization}, self-refinement \cite{madaan2023selfrefineiterativerefinementselffeedback}, and RL-based approaches \cite{deng2022rlpromptoptimizingdiscretetext}.

\paragraph{Paradigm 3: Diagnose.} Our contribution. While Explore methods also iterate and can use error feedback, they treat prompt refinement as a search problem (generate variants, evaluate, select). ERGO instead uses a structured diagnose$\rightarrow$prescribe$\rightarrow$rewrite meta-prompt: at each iteration, it shows the LLM both correct and incorrect predictions from a batch, asks it to identify confused label pairs and why, then jointly rewrites the full prompt (instruction + demonstrations + classification guidelines) in one pass. ProTeGi \cite{pryzant2023automatic} also uses LLM-generated feedback but optimizes instructions only.

\paragraph{Why does this matter?} Concurrent work \cite{diagnosing2025} finds prompt optimization is ``statistically indistinguishable from a coin flip'' across 4 tasks, raising the question of when optimization is worthwhile. Prior work on dataset difficulty \cite{collins2018class} has shown that task characteristics are measurable from data alone; we build on this insight to identify when error-driven refinement specifically provides value. We confirm the overall finding: DSPy's optimization finds no improvement over the default instruction in 35\% of runs, and evaluates candidates on 35-item minibatches that overfit by up to 14 percentage points relative to test accuracy. Error-driven optimization addresses this by focusing search on what actually fails: rather than searching over instruction variants (DSPy/GEPA) or relying on fixed examples (ICL), it learns task-specific conventions directly from structured error diagnosis.

\begin{figure*}[!t]
  \includegraphics[width=\textwidth]{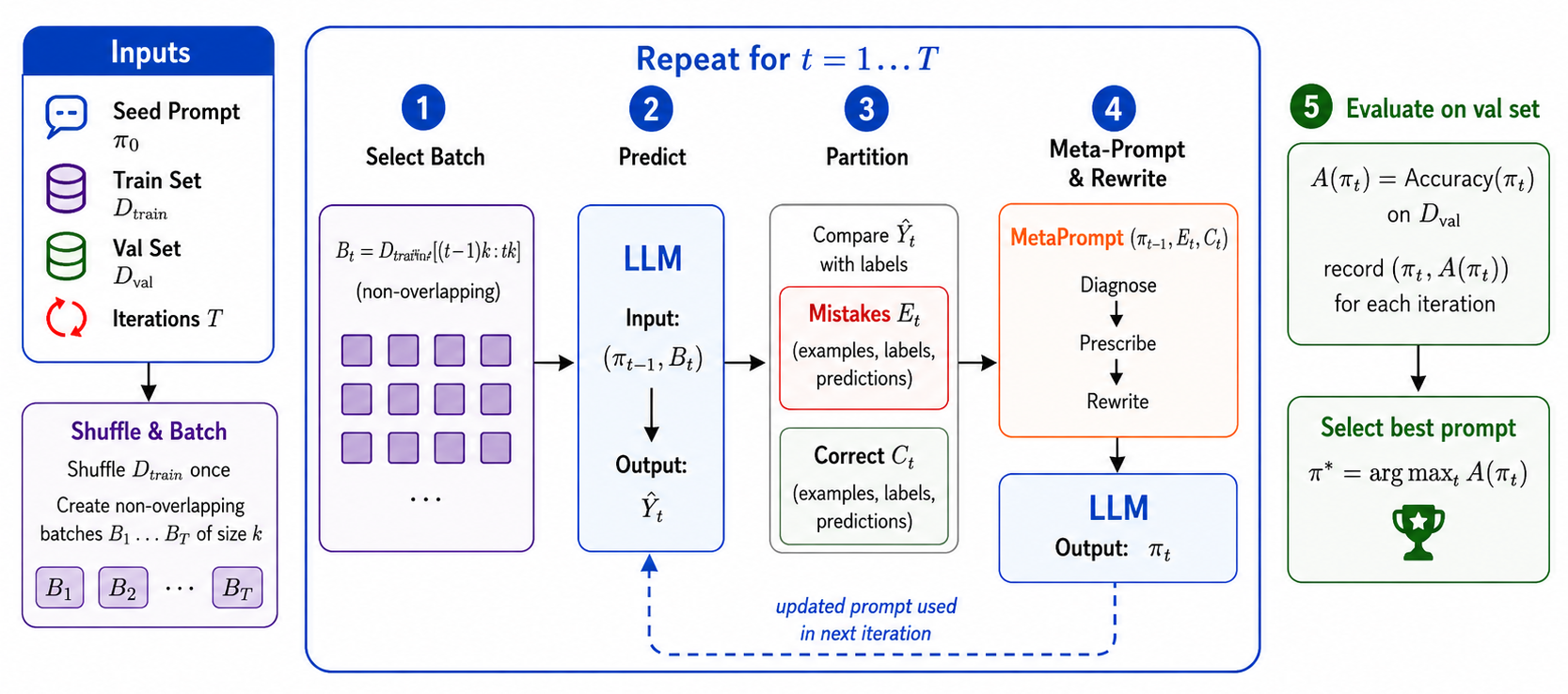}
  \caption{ERGO overview. \textbf{Inputs}: a seed prompt $\pi_0$, training set $D_\text{train}$, validation set $D_\text{val}$, and iteration budget $T$. Training data is shuffled once into non-overlapping batches of size $k$. \textbf{Loop} (steps 1--4): each iteration selects the next batch, classifies it with the current prompt, partitions predictions into mistakes $E_t$ and correct $C_t$, then applies the diagnose$\rightarrow$prescribe$\rightarrow$rewrite meta-prompt to produce an updated prompt $\pi_t$. \textbf{Step 5}: every candidate is evaluated on $D_\text{val}$; the best-on-val prompt $\pi^* = \arg\max_t A(\pi_t)$ is returned.}
  \label{fig:architecture}
\end{figure*}

%%%%%%%%%%%%%%%%%%%%%%%%%%%%%%%%%%%%%%%%%%%%%%%%%%%%%%%%%%%%%%%%%%%%%%%%%%%%%%%
\section{ERGO: Error-Driven Prompt Refinement} \label{sec:method}
%%%%%%%%%%%%%%%%%%%%%%%%%%%%%%%%%%%%%%%%%%%%%%%%%%%%%%%%%%%%%%%%%%%%%%%%%%%%%%%

\subsection{Problem Formulation}

Let $\mathcal{X}$ be an input space and $\mathcal{Y} = \{y_1, \ldots, y_K\}$ a finite label set. We define a prompt $\pi$ as a structured tuple:
\begin{equation}
\pi = (\tau, \mathcal{D}, \mathcal{G})
\end{equation}
where $\tau \in \mathcal{T}$ is a natural language instruction, $\mathcal{D} = \{(x_i, y_i)\}_{i=1}^{n}$ is a set of in-context demonstrations, and $\mathcal{G}$ is a set of decision guidelines (classification rules, boundary conditions, common pitfalls).

Given a frozen LLM $f_\theta : \mathcal{X} \times \Pi \rightarrow \mathcal{Y}$, the prompt optimization problem is:
\begin{align}
\pi^* &= \arg\max_{\pi \in \Pi} \; A(\pi), \\
A(\pi) &= \mathbb{E}_{(x,y) \sim \mathcal{P}} \left[ \mathbf{1}[ f_\theta(x; \pi) = y ] \right]
\end{align}
Since $A(\pi)$ depends on the full tuple $(\tau, \mathcal{D}, \mathcal{G})$ and the components interact (the effectiveness of demonstrations $\mathcal{D}$ depends on the instruction $\tau$ that frames them), we optimize all three jointly rather than in isolation.

\subsection{Overview}

\Cref{fig:architecture} shows the loop. Each iteration selects the next disjoint batch (\textbf{Step 1}), classifies it with the current prompt (\textbf{Step 2}), and partitions predictions into mistakes and correct examples (\textbf{Step 3}). Both go to the meta-prompt, which diagnoses the confused label pairs, prescribes rules resolving them, and rewrites the full prompt (\textbf{Step 4}). The candidate is scored on validation (\textbf{Step 5}); after $T$ iterations the best-scoring candidate is returned. Correct predictions are supplied alongside errors because they constrain the rewrite: without them the model over-generalizes a rule and damages cases the prompt already handled.

We instantiate error-driven refinement as \textbf{Error-Guided Optimization (ERGO)}. Unlike exploration-based search which explores prompt variants, ERGO uses classification errors as structured feedback: at each iteration, it identifies \emph{which} label pairs are confused and generates \emph{why}-level decision rules to disambiguate them. Training examples are partitioned into non-overlapping batches, ensuring every example is seen exactly once across iterations.

\subsection{Algorithm}

\paragraph{Seed prompt.} ERGO starts from a minimal instruction (``Classify the input as exactly one of: \{LABELS\}'') with a structured output format and no demonstrations. The method's gains come entirely from iterative refinement.

\begin{table*}[!t]
  \centering\footnotesize
  \begin{tabular}{l|cc|cccc}
    \toprule
    Dataset & ICL-Uni & ICL-Div & APE & DSPy & GEPA & \textbf{ERGO} \\
    \midrule
    TREC (6 cls) & 87.8 & 89.3 & 86.8 & 84.6 & 85.3 & \textbf{90.0} \\
    CLINC150 (150 cls) & 93.0 & 93.0 & 92.8 & 93.7 & 93.2 & \textbf{94.4} \\
    RTE (2 cls) & 91.8 & 91.3 & 92.5 & 92.7 & 93.0 & \textbf{93.2} \\
    \midrule
    Ethos (2 cls) & 85.6 & 85.4 & 63.4 & \textbf{88.4} & 87.0 & 88.1 \\
    Yahoo (10 cls) & 73.3 & 72.7 & 72.5 & \textbf{73.9} & 72.9 & 71.7 \\
    20Newsgroups (20 cls) & 69.9 & \textbf{70.8} & 67.7 & 69.2 & 68.7 & 66.8 \\
    Rotten Tom. (2 cls) & 93.2 & 93.2 & 92.6 & 92.9 & \textbf{93.7} & 92.8 \\
    MASSIVE (60 cls) & \textbf{90.1} & 89.2 & 89.3 & 90.0 & 89.4 & 87.8 \\
    \midrule
    \textbf{Average (8)} & 85.6 & 85.6 & 82.2 & \textbf{85.7} & 85.4 & 85.6 \\
    \bottomrule
  \end{tabular}
  \caption{Test accuracy (\%) on 8 classification benchmarks (mean over 5 seeds; $\pm$std in \cref{tab:results-std}). \textbf{Bold} = best. No method significantly outperforms all others overall (paired $t$-test, $p{>}0.05$).}
  \label{tab:results}
\end{table*}

\paragraph{Feedback-first meta-prompt.} At each iteration, the meta-prompt provides the LLM with: (a) all correctly classified examples from the batch, (b) all misclassified examples with predicted vs.\ true labels, and instructs the LLM to: (1) \emph{diagnose} which label pairs are confused and why, (2) \emph{prescribe} specific decision rules, (3) \emph{rewrite} the full prompt (full template in \cref{sec:appendix-metaprompt}; pseudocode in \cref{alg:ergo}). A key byproduct is interpretability: the resulting prompts contain human-readable decision rules that practitioners can audit, refine, or use as annotation guidelines.

\paragraph{Random non-overlapping batches.} Training examples are shuffled once; sequential 20-item batches ensure each example is seen exactly once across iterations, maximizing information coverage without stratification complexity.

\paragraph{Best-of-$T$ selection.} Since LLM generation is stochastic and improvement is not guaranteed at every step, ERGO retains all candidate prompts and returns the best by validation accuracy. Each candidate is evaluated on the full validation set to reduce selection noise. Empirically, the optimal prompt is found within 3--5 iterations.

%%%%%%%%%%%%%%%%%%%%%%%%%%%%%%%%%%%%%%%%%%%%%%%%%%%%%%%%%%%%%%%%%%%%%%%%%%%%%%%
\section{Experiments} \label{sec:experiments}

%%%%%%%%%%%%%%%%%%%%%%%%%%%%%%%%%%%%%%%%%%%%%%%%%%%%%%%%%%%%%%%%%%%%%%%%%%%%%%%

\subsection{Setup}

\paragraph{Datasets.} We evaluate on 8 representative classification benchmarks spanning 2--150 classes, including Ethos \cite{mollas2006ethos}, RTE \cite{wang2018glue}, and CLINC150 among others (full details in \cref{sec:appendix-datasets}). Each uses a stratified split targeting 200 training, 100 validation, and 500 test examples, reduced proportionally for smaller datasets (CLINC150: 150 validation examples, 1/class with full label coverage).

\paragraph{Model.} Claude Sonnet 4.5 via AWS Bedrock (temperature=0.0). Per run, ERGO completes in 16.4 min on average vs.\ DSPy (28.8 min) and GEPA (30.9 min); full cost comparison in \cref{tab:cost}. All hyperparameters are consolidated in \cref{sec:appendix-hyper}.

\paragraph{Seeds.} All experiments use 5 random seeds (42, 123, 7, 1, 2). Statistical significance via paired $t$-test at 5\% level across all 40 configurations (8 datasets $\times$ 5 seeds).

\paragraph{Baselines.} (1) ICL-Uniform: 20 random demos; (2) ICL-Diversity: 20 demos via $k$-means on Sentence-BERT \cite{reimers2019sentence}; (3) APE \cite{zhou2023largelanguagemodelshumanlevel}: best-of-5 instructions; (4) DSPy \cite{khattab2023dspycompilingdeclarativelanguage,sarmah2024comparativestudydspyteleprompter}: MIPROv2 \texttt{auto="light"}; (5) GEPA \cite{agrawal2026gepa}: evolutionary prompt optimization (our primary comparison target as the closest iterative method). We also evaluate ProTeGi \cite{pryzant2023automatic}, another error-feedback method; its $7.7\times$ higher compute cost and instruction-only optimization make it less directly comparable, so we report it separately in \cref{sec:appendix-protegi}.

\subsection{Main Results}

\Cref{tab:results} shows results on 8 classification benchmarks.

\paragraph{Error structure identifies where optimization pays off.} On tasks whose errors concentrate in a learnable boundary, ERGO discovers reusable decision rules. On TREC, one confused pair (entities $\leftrightarrow$ human beings) accounts for most errors and one rule resolves them: this is a \textbf{pair-local} boundary, and ERGO reaches 90.0\% (+4.7 pp over GEPA). CLINC150 exhibits \textbf{shared structural} boundaries, where several rare pairs each express the same distinction (a specific intent collapsing into its general sibling), and ERGO leads with 94.4\% (+1.2\%, $p{=}0.043$). In both cases the errors satisfy the three conditions of boundary learnability: a recurring distinction surfaces (\emph{discovery}), it compresses into reusable rules (\emph{structure}), and applying the rules improves accuracy with limited collateral damage (\emph{repairability}).
\paragraph{Explore and Demonstrate win on coverage tasks.} ICL-Diversity wins on 20Newsgroups (70.8\% vs ERGO's 66.8\%, $p{=}0.013$), ICL-Uniform on MASSIVE (90.1\% vs 87.8\%, $p{<}0.001$), and DSPy on Yahoo (73.9\% vs 71.7\%, $p{=}0.046$). These tasks require broad coverage of many overlapping categories whose errors do not concentrate in a few learnable pairs, so demonstrations or exploration outperform targeted rules (\cref{tab:main-error-structure} in the appendix summarizes the error structure of all benchmarks).
\paragraph{Overall: no method dominates.} ERGO averages 85.6\%, not significantly different from ICL-Diversity (85.6\%), DSPy (85.7\%), or GEPA (85.4\%), all pairwise $p{>}0.6$. Only APE is significantly worse ($p{=}0.013$). The ordering is unchanged under macro-F1 and macro-recall (\cref{sec:appendix-macro}). The contribution is not a new state-of-the-art but identifying when each paradigm provides unique value: ERGO for boundary-learnable tasks, ICL for coverage tasks, and DSPy/GEPA for broad-coverage optimization.

%%%%%%%%%%%%%%%%%%%%%%%%%%%%%%%%%%%%%%%%%%%%%%%%%%%%%%%%%%%%%%%%%%%%%%%%%%%%%%%
\section{Quantitative Analysis} \label{sec:case_studies}

\subsection{Convergence: Rapid Error-Driven Learning}

\begin{figure*}[t]
  \centering
  \begin{subfigure}[b]{0.48\textwidth}
    \includegraphics[width=\textwidth]{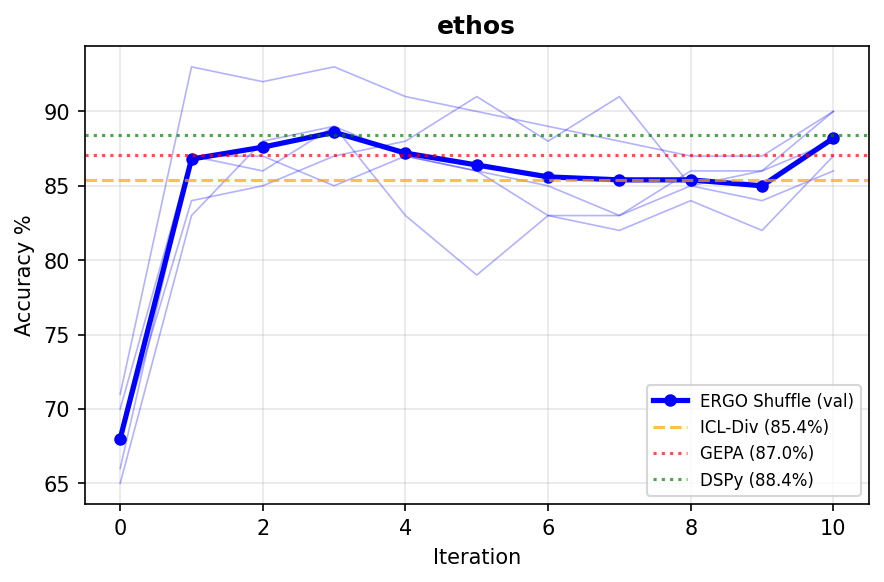}
    \caption{Ethos: 68\%$\rightarrow$88.6\% in 3 iters}
  \end{subfigure}
  \hfill
  \begin{subfigure}[b]{0.48\textwidth}
    \includegraphics[width=\textwidth]{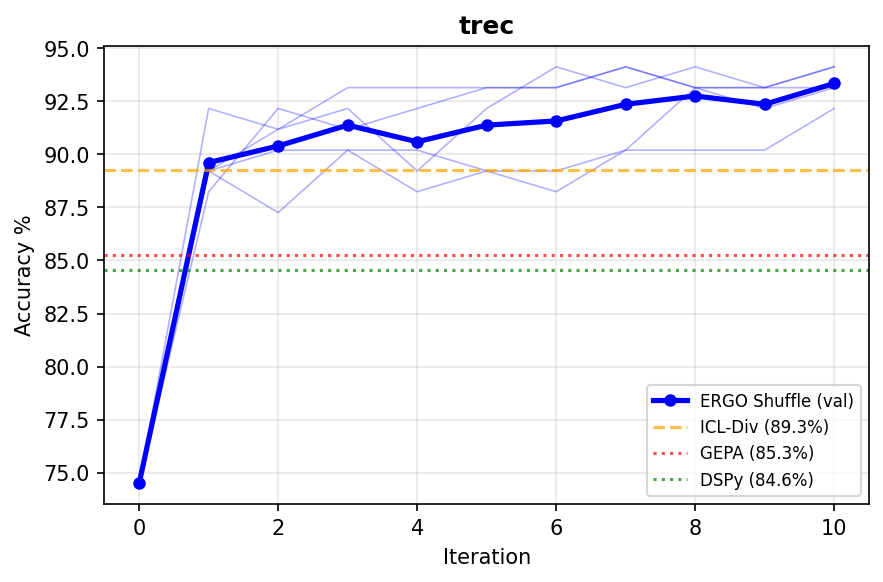}
    \caption{TREC: 74.5\%$\rightarrow$91.4\% in 3 iters}
  \end{subfigure}
  \caption{ERGO convergence on boundary-learnable tasks. Dashed lines: ICL-Div, GEPA, and DSPy baselines. ERGO captures 80\% of total gain in the first 3 iterations.}
  \label{fig:convergence}
\end{figure*}

\Cref{fig:convergence} shows ERGO's convergence on the two tasks with most dramatic early improvement. On Ethos, ERGO jumps from 68\% to 88.6\% (+20.6\%) in just 3 iterations as it learns the profanity-vs-targeting boundary. On TREC, it jumps from 74.5\% to 91.4\% (+16.9\%) as it discovers the organizations-as-human-beings convention. Both surpass GEPA's final accuracy within 1--2 iterations; on TREC, ERGO also surpasses DSPy. Rapid convergence is a signature of boundary learnability: when errors have \textbf{pair-local} structure, a single iteration can discover the dominant confused pair and prescribe a rule that repairs it, so most of the gain is available early.

%\subsection{Cross-Model Transfer}
%
%ERGO is the best \emph{optimization} method for cross-model deployment (\cref{tab:transfer} in Appendix). While ICL transfers slightly better (it carries demonstrations any model can use), among methods that optimize prompts, ERGO significantly outperforms DSPy ($p{=}0.035$, $+$6.5pp) on Haiku. The key advantage is robustness: ERGO's decision rules never catastrophically fail on a weaker model, while DSPy's demonstrations cause catastrophic failures (CLINC150: 93.7\%$\rightarrow$44.2\%).

\subsection{Homogeneous Training: Model Generalization}
\label{sec:gen-homo}

To test whether ERGO generalizes beyond Claude, we run the full protocol (optimize, validate, and infer with the same model) on five models from four families: Claude Haiku 4.5 and Sonnet 4.5, Gemma 3 27B IT, and Mistral Large 3 (\cref{tab:gen-homo}; per-dataset breakdown in \cref{tab:gen-homo-full}). Pooled across all 200 (model, dataset, seed) triples, ERGO improves over the baseline by $+5.67$ pp ($t{=}6.34$, $p{=}1.5{\times}10^{-9}$). Four of five models gain significantly with the same meta-prompt and no model-specific tuning. ERGO wins on $61.0\%$ of individual (dataset, seed) pairs.

\paragraph{Largest gains come from weak-baseline datasets.} ETHOS ($+28.7$ pp) and TREC ($+16.4$ pp) drive most of the cross-model improvement: the seed prompt produces near-random output for several models, and ERGO's error-driven feedback corrects these issues within a few iterations. On near-ceiling tasks ($\geq 0.93$ baseline) gains are minimal.

\begin{table}[t]
  \centering\footnotesize
  \begin{tabular}{l|cc|c}
    \toprule
    Model & Baseline & ERGO & $\Delta$ (pp) \\
    \midrule
    Sonnet 4.5 & 80.30 & \textbf{85.55} & $+5.25^{**}$ \\
    Haiku 4.5 & 74.09 & 82.37 & $+8.29^{**}$ \\
    Mistral L3 & 75.44 & 78.97 & $+3.53^{*}$ \\
    Gemma 3 27B & 69.33 & 70.51 & $+1.18^{\text{n.s.}}$ \\
    \midrule
    \textbf{Overall} & \textbf{74.31} & \textbf{79.98} & $\mathbf{+5.67^{***}}$ \\
    \bottomrule
  \end{tabular}
  \caption{Homogeneous training: ERGO vs.\ seed-prompt baseline across model families (8 datasets, 5 seeds, $N{=}40$ per model). $\Delta$ in percentage points; $p$-value from paired $t$-test. $^{***}\,p{<}0.001$, $^{**}\,p{<}0.01$, $^{*}\,p{<}0.05$.}
  \label{tab:gen-homo}
\end{table}

\subsection{Homogeneous Training: Ablation Studies}

We isolate the effect of four ERGO hyperparameters on Sonnet 4.5 across all 8 datasets with 3 seeds (\cref{tab:ablation} in Appendix). Across all 16 variants the spread is only $1.40$ pp (84.85--86.24), and no variant differs significantly from the default except $k{=}50$ batch size ($+1.29$ pp, $p{=}7.5{\times}10^{-3}$). Key findings: accuracy plateaus by T${=}5$ (99\% of T${=}10$ at 73\% cost), shuffle outperforms class-balanced sampling while using 18\% fewer calls, and joint rewriting of all components avoids catastrophic failures on boundary-learnable tasks despite demos-only yielding marginally higher average accuracy. ERGO is robust to hyperparameter choice.

\subsection{Heterogeneous Training: Sonnet Refines, Haiku Infers}
\label{sec:gen-hetero}

We next decouple the refinement and inference models. Claude Sonnet 4.5 runs the ERGO loop (generating refined prompts from its own classifications and feedback), and the resulting best-on-val prompt is then used by Claude Haiku 4.5 for test-time inference (\cref{tab:gen-hetero}; per-dataset numbers in \cref{tab:gen-hetero-full}). Sonnet-refined prompts give Haiku a $+8.83$ pp lift over its seed-prompt baseline ($t{=}3.52$, $p{=}1.1{\times}10^{-3}$). The gap to homogeneous Haiku-ERGO is $+0.55$ pp, which is not significant ($p{=}0.22$). The optimized prompt thus carries task knowledge that transfers cleanly to a smaller inference model: a stronger model can be used once for refinement, and the resulting prompt can be deployed with a cheaper model without statistically significant accuracy loss ($p{=}0.22$).

\begin{table}[t]
  \centering\footnotesize
  \begin{tabular}{l|c}
    \toprule
    Setting & Acc \\
    \midrule
    Baseline (seed prompt, Haiku) & 74.1 \\
    Heterogeneous ERGO (Sonnet $\rightarrow$ Haiku) & \textbf{82.9} \\
    Homogeneous ERGO (Haiku $\rightarrow$ Haiku) & 82.4 \\
    \bottomrule
  \end{tabular}
  \caption{Heterogeneous training: prompts refined on Sonnet 4.5 are used by Haiku 4.5 at test time (8 datasets, 5 seeds). Hetero vs.\ baseline: $p{=}0.001$; hetero vs.\ homo: $p{=}0.22$ (n.s.).}
  \label{tab:gen-hetero}
\end{table}

\subsection{Robustness to Annotation Noise}

We identify 162 queries (4.6\% of all test data) where ERGO, DSPy, and GEPA across all 5 seeds unanimously predict the same ``wrong'' answer. Manual review confirms the majority are clear annotation errors (e.g., gold labels contradicting the text), with the remainder being ambiguous boundary cases. Removing these, the relative performance gaps are unchanged (ERGO vs GEPA: +0.3\% raw $\rightarrow$ +0.3\% clean), confirming that ERGO's advantage comes from learning genuine task conventions, not fitting annotation noise.

\begin{table*}[!t]
  \centering\footnotesize
  \begin{tabular}{l|l|l|l}
    \toprule
    Task characteristic & Recommended paradigm & Method & Example datasets \\
    \midrule
    Learnable boundaries & Diagnose & ERGO & TREC, Ethos, CLINC150 \\
    Need interpretable output & Diagnose & ERGO & Ethos \\
    Many-class + broad & Explore & DSPy & MASSIVE, Yahoo \\
    Many overlapping classes & Demonstrate & ICL & 20News \\
    Zero optimization budget & Demonstrate & ICL & Any \\
    Near-ceiling tasks & Any & ICL (cheapest) & Rotten Tom. \\
    \bottomrule
  \end{tabular}
  \caption{Paradigm selection guide based on task characteristics.}
  \label{tab:when}
\end{table*}

\section{Case Studies} \label{sec:case_studies_qual}

The quantitative results show that ERGO wins on boundary-learnable tasks; here we show why, with each case study illustrating a different condition of boundary learnability: \emph{discovery} of a recurring distinction from a single error (TREC), \emph{structure} where one rule repairs an entire class of confusions (CLINC150), and \emph{repairability} through progressive refinement (Ethos).

\subsection{TREC: Learning Hidden Conventions}

TREC (6-class question type classification) illustrates \emph{pair-local} boundary learnability: errors concentrate in one confused pair, and a single rule resolves them. The task contains a counterintuitive labeling convention: questions about companies, schools, or organizations should be classified as ``human beings,'' a rule that cannot be inferred from the label name alone. For instance, ``What is the first personal computer company?'' and ``What army's motto is Blood and Fire?'' are both gold-labeled ``human beings'' (full table in \cref{sec:appendix-cases}).

ERGO discovers this convention at iteration 3 (seed 123), when the batch contains ``What Japanese electronics company was named for a coastal city?'' misclassified as ``entities.'' The meta-prompt diagnoses: \emph{``The classifier incorrectly treats companies as entities rather than human organizations,''} and prescribes: \emph{``Add rule: `What [company/organization/band/team]' $\rightarrow$ human beings.''} This single error triggers a rule that fixes 25+ test queries across all seeds. Per-class analysis confirms ERGO's gains concentrate on the ``human beings'' class (+5.5 pp over ICL-Diversity; \cref{tab:perclass}).

\subsection{CLINC150: Structural Rule Generalization}

CLINC150 (150-class intent detection) illustrates \emph{shared structural} boundaries: many individually rare confused pairs all instantiate the same distinction, so a single rule repairs them together. The task contains many near-identical label pairs that differ only by action vs.\ status: ``set up a reminder so I don't forget the baby shower'' should map to \texttt{reminder\_update} (creating), not \texttt{reminder} (querying); similarly, ``how did my bank account get frozen'' maps to \texttt{account\_blocked} (status), not \texttt{freeze\_account} (action).

ERGO learns the general principle from a single misclassified example: creating or modifying maps to \texttt{\_update} variants, while querying or viewing maps to base variants. This structural rule generalizes across multiple label pairs (reminder/reminder\_update, calendar/calendar\_update, freeze\_account/account\_blocked): a single principle learned from one error fixes an entire class of confusions (\cref{tab:clinc_case}).

\subsection{Ethos: Progressive Boundary Refinement}

Ethos (binary hate speech detection) illustrates \emph{repairability}: each iteration prescribes a rule that repairs a subset of errors without damaging cases the prompt already handles. ERGO learns the profanity-vs-targeting boundary progressively over 3 iterations, jumping from 68\% to 86.8\% at iteration 1 by prescribing ``True requires BOTH targeting based on identity AND dehumanization/threats,'' then refining with two further rules about abstract dismissals and non-protected characteristics (full progression in \cref{sec:appendix-cases}). The final prompt is directly interpretable and auditable as an annotation guideline. We note that this rule captures the annotation criterion used in the Ethos dataset; other hate-speech datasets with different annotation guidelines may require different boundary rules, which ERGO would discover from their respective error patterns.

\section{Comparative Analysis} \label{sec:complementarity}

The three paradigms are complementary: \emph{demonstrate} (ICL) excels on coverage-dependent tasks, \emph{explore} (DSPy/GEPA) on many-class tasks, and \emph{diagnose} (ERGO) on boundary-learnable tasks.

\subsection{ERGO vs.\ GEPA: Error-Driven vs.\ Exploration-Based}

We compare against GEPA as the closest iterative method: both refine instruction-based prompts, but GEPA generates rules via evolutionary search while ERGO learns them incrementally from errors.

Across our benchmarks, we identify 113 queries where ERGO consistently gets right but GEPA gets wrong (3+ seeds), and 84 in the reverse direction.

\paragraph{Why they fail in opposite ways:} ERGO over-constrains (rules too specific misfire on edge cases); GEPA under-constrains (generic guidelines miss task-specific boundaries). On TREC, GEPA's rule covers only organizations that ``did something,'' missing identification questions. Conversely, ERGO's ``profanity $\neq$ hate'' rule misses gendered slurs that GEPA's broader definition catches. The failure modes reflect boundary structure: ERGO excels on \emph{pair-local} and \emph{shared structural} tasks where errors compress into reusable rules (TREC, Ethos, CLINC150); exploration and demonstrations win on coverage tasks (20Newsgroups, Yahoo, MASSIVE) where errors do not concentrate in learnable pairs (see \cref{sec:appendix-20news} for a detailed 20Newsgroups example).

\paragraph{Convergence and stability:} On boundary-learnable tasks, ERGO converges in 3 iterations (Ethos: 68\%$\rightarrow$88.6\%, TREC: 74.5\%$\rightarrow$91.4\%), surpassing GEPA's final accuracy within 1--2 iterations. GEPA finds one good mutation early then cannot improve further (Ethos: 1 accepted out of 124 iterations). ERGO is 4--10$\times$ more stable across seeds (Ethos: std 0.13\% vs 1.25\%) because learned rules apply deterministically (\cref{tab:results-std}).

% \paragraph{Prompt characteristics:} ERGO prompts grow from 36 words to 600--1,350 words over 8--11 iterations of error-driven refinement. GEPA generates 150--876 word prompts in one shot. ERGO's prompts are longer but contain explicit, auditable decision rules; GEPA's are shorter but rely on the LLM's implicit reasoning. Notably, DSPy selects zero demonstrations in 46\% of runs despite its joint instruction+demo optimization design, indicating that its selected demos can introduce noise that hurts classification on certain tasks.

\subsection{ERGO vs.\ ICL: Rules vs.\ Demonstrations}

ICL-Diversity edges ERGO overall (85.9\% vs.\ 85.6\%), though the difference is not statistically significant. ERGO wins on 4 boundary-learnable tasks (Ethos +2.6\%, RTE +1.9\%, CLINC150 +1.4\%, TREC +0.7\%); ICL wins on broad-coverage tasks (20Newsgroups $-$4.0\%, MASSIVE $-$1.3\%).

\paragraph{Why they differ:} Demonstrations cannot express abstract rules. On Ethos, a demo showing ``Kill them all = False'' helps one query, but ERGO's explicit rule ``profanity without group targeting = False'' generalizes to all similar cases. On CLINC150, 20 demos cover only 13\% of 150 labels and cannot demonstrate the action-vs-status distinction for most pairs. Conversely, on broad-coverage tasks, ICL's holistic pattern matching outperforms ERGO's keyword-based rules that misfire across overlapping categories. The fundamental tradeoff: rules are precise but brittle; demonstrations are robust but implicit.

\subsection{When Does Each Paradigm Win?}

A key finding is that the cheapest method (ICL-Diversity, zero optimization cost) is already optimal on the majority of tasks — iterative optimization is justified only under specific, identifiable conditions (\cref{tab:when}).

We recommend the following decision procedure: (1) start with ICL-Diversity as a zero-cost baseline; (2) inspect the errors: if they reveal a recurring distinction (\emph{discovery}) that is either \emph{pair-local} (one dominant confused pair) or \emph{shared structural} (many rare pairs expressing the same distinction), apply ERGO; verify the rules help without collateral damage (\emph{repairability}); (3) if errors spread across many pairs with no shared mechanism, try DSPy's structured taxonomy framing; (4) when interpretability or annotation guidelines are needed, prefer ERGO regardless of margin.

For near-ceiling tasks where all methods converge to similar accuracy, ICL is the cheapest option and optimization provides no statistically significant benefit. We validate this procedure on 6 additional datasets (\cref{sec:appendix-held-out}): errors spread across many pairs with no shared mechanism reliably identify tasks where optimization is unnecessary, and concentrated errors with \emph{pair-local} or \emph{shared structural} form that prove repairable identify tasks where ERGO provides value.

%%%%%%%%%%%%%%%%%%%%%%%%%%%%%%%%%%%%%%%%%%%%%%%%%%%%%%%%%%%%%%%%%%%%%%%%%%%%%%%
\section{Conclusion}
%%%%%%%%%%%%%%%%%%%%%%%%%%%%%%%%%%%%%%%%%%%%%%%%%%%%%%%%%%%%%%%%%%%%%%%%%%%%%%%

We propose ERGO, a prompt optimization method that iterates over the full training set in non-overlapping batches, diagnoses classification errors using both mistakes and correct predictions as contrast, and generates interpretable decision rules. Through case studies on 8 classification benchmarks, we demonstrate that error-driven refinement discovers task conventions invisible to other paradigms, such as ``organizations = human beings'' in TREC and ``profanity without targeting $\neq$ hate speech'' in Ethos.

Our central finding is that the three paradigms (demonstrate, explore, and diagnose) are complementary rather than competing: no method significantly outperforms the others overall. ERGO excels on boundary-learnable tasks, those whose errors satisfy \emph{discovery}, \emph{structure}, and \emph{repairability}, while ICL-Diversity on coverage-dependent tasks at zero optimization cost, and DSPy/GEPA on many-class tasks requiring broad exploration.

The practical contribution is a paradigm selection framework linking task characteristics to the optimal method. Future work includes adapting batch strategies for many-class settings and exploring hybrid approaches that combine multiple paradigms.

%%%%%%%%%%%%%%%%%%%%%%%%%%%%%%%%%%%%%%%%%%%%%%%%%%%%%%%%%%%%%%%%%%%%%%%%%%%%%%%
\section*{Limitations}
%%%%%%%%%%%%%%%%%%%%%%%%%%%%%%%%%%%%%%%%%%%%%%%%%%%%%%%%%%%%%%%%%%%%%%%%%%%%%%%

\paragraph{Not universally better.} ERGO is not significantly different from ICL-Diversity overall, though it shows lower accuracy on coverage-dependent tasks such as 20Newsgroups and MASSIVE. We do not claim dominance; the contribution is identifying when error-driven refinement provides unique value.

\paragraph{Task scope.} ERGO is designed for classification with learnable boundaries and is not suitable for generation or multi-hop reasoning. All benchmarks are English-only, and while they span diverse domains and class counts, many task domains (e.g., legal, medical) remain untested; generalizability beyond our benchmark suite remains to be validated.

\paragraph{Scaling and batch coverage.} The fixed batch size ($k{=}20$) may need adjustment for many-class tasks. On CLINC150 (150 classes) and MASSIVE (60 classes), random 20-item batches cover only 10--15\% of classes per iteration, limiting error signal diversity. As the number of classes grows, the generated rules may not generalize across all class boundaries, and limited or imbalanced training data may not expose representative error patterns.

\paragraph{Heterogeneous training assumption.} We assume the learning model (which generates rules) is at least as strong as the task model (which applies them at inference). Weak-to-strong settings (e.g., Haiku optimizing prompts for Sonnet) remain to be validated.

\paragraph{Use of AI assistants.} AI writing assistants were used for language editing and literature searching during the preparation of this manuscript. All generated content was verified, revised, and approved by the authors, who take full responsibility for the final text.

\bibliography{reference}

%%%%%%%%%%%%%%%%%%%%%%%%%%%%%%%%%%%%%%%%%%%%%%%%%%%%%%%%%%%%%%%%%%%%%%%%%%%%%%%
\appendix
%%%%%%%%%%%%%%%%%%%%%%%%%%%%%%%%%%%%%%%%%%%%%%%%%%%%%%%%%%%%%%%%%%%%%%%%%%%%%%%

\section{Dataset Details} \label{sec:appendix-datasets}

\Cref{tab:datasets} summarizes the 8 classification benchmarks used in our evaluation, spanning 2 to 150 classes across diverse domains. Training sets provide ERGO's refinement batches. Validation sets are used for prompt selection. Test sets are held out for final reporting. All datasets are publicly available and used consistent with their intended purpose of classification benchmarking. No dataset is used outside its original research context.

\begin{table*}[!htbp]
  \centering\footnotesize
  \begin{tabular}{l|r|rrr|l}
    \toprule
    Dataset & Classes & Train & Val & Test & Domain \\
    \midrule
    TREC \cite{li2002learning} & 6 & 198 & 102 & 451 & Question type \\
    CLINC150 \cite{larson2019evaluation} & 150 & 150 & 150 & 450 & Intent detection \\
    RTE \cite{wang2018glue} & 2 & 80 & 40 & 157 & NLI \\
    Ethos \cite{mollas2006ethos} & 2 & 200 & 100 & 698 & Hate speech \\
    Yahoo Answers \cite{NIPS2015_250cf8b5} & 10 & 200 & 100 & 500 & Topic \\
    Rotten Tomatoes \cite{leone2020rottentomatoes} & 2 & 200 & 100 & 500 & Sentiment \\
    20Newsgroups \cite{lang1995newsweeder} & 20 & 200 & 100 & 500 & Topic \\
    MASSIVE \cite{fitzgerald2022massive} & 60 & 120 & 60 & 288 & VA intent \\
    \bottomrule
  \end{tabular}
  \caption{Dataset statistics. All splits are stratified; many-class datasets ensure $\geq$2 examples per class in training.}
  \label{tab:datasets}
\end{table*}

\section{Case Study Details} \label{sec:appendix-cases}

\subsection{TREC: Query Examples}
\begin{table}[H]
  \centering\footnotesize
  \begin{tabular}{p{5.5cm}|c}
    \toprule
    Query & Gold \\
    \midrule
    ``What is the first personal computer company?'' & human beings \\
    ``What army's motto is Blood and Fire?'' & human beings \\
    ``Rotary engine cars were made by what company?'' & human beings \\
    ``What group asked `Do You Believe in Magic'?'' & human beings \\
    ``What party was Winston Churchill a member of?'' & human beings \\
    ``What are the major companies that are part of Dow Jones?'' & human beings \\
    \bottomrule
  \end{tabular}
  \caption{TREC queries: ERGO correct (5/5 seeds), GEPA wrong (5/5 seeds). GEPA predicts ``entities'' for all.}
  \label{tab:trec_case}
\end{table}

\begin{table}[H]
  \centering\footnotesize
  \begin{tabular}{l|ccc}
    \toprule
    Class & ICL-Div & ERGO & $\Delta$ \\
    \midrule
    human beings & 89.9\% & 95.4\% & \textbf{+5.5} \\
    description & 89.2\% & 92.8\% & $+$3.6 \\
    entities & 81.2\% & 82.2\% & $+$1.0 \\
    numeric values & 99.8\% & 98.8\% & $-$1.0 \\
    locations & 85.5\% & 82.2\% & $-$3.4 \\
    abbreviation & 91.1\% & 86.7\% & $-$4.4 \\
    \bottomrule
  \end{tabular}
  \caption{Per-class accuracy on TREC: ERGO's gains concentrate on the target class (human beings).}
  \label{tab:perclass}
\end{table}

\subsection{Ethos: Query Examples}

ERGO learns the profanity-vs-targeting boundary progressively. At iteration 1 (9 mistakes), it prescribes: ``True requires BOTH targeting based on identity AND dehumanization/threats,'' jumping from 68\% to 86.8\%. Iteration 2 (6 mistakes) adds: ``dismissing a condition in abstract terms $\neq$ targeting people with that condition.'' Iteration 3 (4 mistakes) adds: ``economic status and occupation are NOT protected characteristics.'' The result: texts with profanity but no identity-group targeting are correctly classified as False.

\begin{table}[H]
  \centering\footnotesize
  \begin{tabular}{p{6cm}|c}
    \toprule
    Text & Seeds \\
    \midrule
    ``Holy shit! Kill them all!'' & 4/5 \\
    ``Go f***ing kill yourself... useless pile of shit scumbag'' & 4/5 \\
    ``Not only the politicians are going into the wood chippers but also the police'' & 4/5 \\
    ``Every time that blonde lady talks I lose twenty brain cells'' & 3/5 \\
    ``I very much appreciate the Asian woman's smarts'' & 3/5 \\
    \bottomrule
  \end{tabular}
  \caption{Ethos: ERGO correct, GEPA wrong (4--5/5 seeds). Gold = False (not hate speech). GEPA predicts True for all.}
  \label{tab:ethos_case}
\end{table}

\subsection{CLINC150: Query Examples}
\begin{table*}[!htbp]
  \centering\footnotesize
  \begin{tabular}{p{6cm}|c|c}
    \toprule
    Query & Gold & GEPA pred \\
    \midrule
    ``set up a reminder so I don't forget the baby shower'' & reminder\_update & reminder \\
    ``i need to get a new credit card application'' & new\_card & application\_status \\
    ``how long will I have to wait at cheesecake factory'' & how\_busy & restaurant\_reviews \\
    ``what is the best way to remove this appointment'' & calendar\_update & cancel\_reservation \\
    \bottomrule
  \end{tabular}
  \caption{CLINC150: ERGO correct, GEPA wrong (4/5 seeds).}
  \label{tab:clinc_case}
\end{table*}

ERGO learns that intent labels distinguish between \emph{performing an action} and \emph{querying status}:

\begin{itemize}[noitemsep, left=0pt]
\item ``set up a reminder so I don't forget the baby shower'' $\rightarrow$ \texttt{reminder\_update} (creating), not \texttt{reminder} (querying)
\item ``how did my bank account get frozen'' $\rightarrow$ \texttt{account\_blocked} (status), not \texttt{freeze\_account} (action)
\item ``what is the best way to remove this appointment'' $\rightarrow$ \texttt{calendar\_update} (modifying), not \texttt{cancel\_reservation}
\end{itemize}

GEPA's generic prompt cannot teach this structural distinction. ICL's 20 demos (covering only 13\% of 150 labels) are unlikely to include both members of these confusable pairs.

\subsection{Where GEPA Wins: 20Newsgroups} \label{sec:appendix-20news}

GEPA outperforms ERGO by 1.9\% on 20Newsgroups because its evolutionary search produces holistic guidelines: \emph{``Look for context beyond surface-level keywords. A discussion about `scientific facts' may actually be part of a religious debate.''} ERGO's keyword-based rules (learned from 20-item batches) cannot capture this holistic understanding of 20 overlapping categories.

\subsection{Rotten Tomatoes: Conditional Recommendations}

ERGO learns that hedged or conditional positive language does not override strong negative criticism:

\begin{itemize}[noitemsep, left=0pt]
\item ``the four feathers is definitely horse feathers, but if you go in knowing that, you might have fun'' (Gold: Negative, initial: Positive)
\item ``moderately involving despite bargain-basement photography and hackneyed romance'' (Gold: Negative, initial: Positive)
\end{itemize}

ERGO's rule: \emph{``Weak conditional positives (`you might have fun if...') do not override strong negative qualifiers (`bargain-basement,' `hackneyed'). Classify based on the dominant critical assessment.''}

\section{Cross-Model Transfer and Generalization} \label{sec:appendix-transfer}

%\begin{table}[H]
%  \centering\footnotesize
%  \begin{tabular}{l|cc|c}
%    \toprule
%    Method & Sonnet & Haiku & Retention \\
%    \midrule
%    ICL-Diversity & 85.6\% & 84.1\% & 98.2\% \\
%    \textbf{ERGO} & 85.6\% & 83.4\% & 97.4\% \\
%    GEPA & 85.4\% & 83.0\% & 97.2\% \\
%    APE & 82.2\% & 75.9\% & 92.4\% \\
%    DSPy & 85.7\% & 76.9\% & 89.8\% \\
%    \bottomrule
%  \end{tabular}
%  \caption{Cross-model transfer: prompts optimized on Sonnet 4.5, evaluated on Haiku 4.5 (8 datasets, 5 seeds). Retention = Haiku accuracy / Sonnet accuracy (\%).}
%  \label{tab:transfer}
%\end{table}

\begin{table*}[!htbp]
    \centering\footnotesize
    \begin{tabular}{l|cc|cc|cc|cc}
      \toprule
      & \multicolumn{2}{c|}{Haiku 4.5} & \multicolumn{2}{c|}{Sonnet 4.5} & \multicolumn{2}{c|}{Gemma 3 27B} & \multicolumn{2}{c}{Mistral L3} \\
      Dataset & B & I & B & I & B & I & B & I \\
      \midrule
      TREC & 51.4 & \textbf{84.7} & 71.9 & \textbf{90.0} & \textbf{41.5} & 41.4 & 58.6 & \textbf{60.0} \\
      CLINC150 & 90.2 & \textbf{92.8} & 92.7 & \textbf{94.2} & \textbf{85.6} & 85.4 & 88.4 & \textbf{88.7} \\
      RTE & 91.7 & \textbf{93.5} & 93.1 & \textbf{93.6} & 86.6 & 86.6 & 88.8 & \textbf{89.7} \\
      Ethos & 48.9 & \textbf{86.2} & 63.4 & \textbf{87.7} & 42.3 & \textbf{50.4} & 54.8 & \textbf{83.4} \\
      Yahoo & \textbf{71.6} & 66.2 & \textbf{72.8} & 71.2 & 70.8 & 70.8 & \textbf{73.3} & 71.4 \\
      Rotten Tom. & \textbf{91.6} & 89.9 & 91.4 & \textbf{93.0} & 90.2 & \textbf{91.9} & 89.4 & \textbf{90.4} \\
      20News & \textbf{63.0} & 62.6 & \textbf{67.6} & 67.0 & 57.5 & \textbf{57.6} & \textbf{63.7} & 63.0 \\
      MASSIVE & \textbf{84.2} & 83.1 & \textbf{89.4} & 87.7 & \textbf{80.1} & 79.9 & \textbf{86.7} & 85.2 \\
      \midrule
      \textbf{Average} & 74.1 & \textbf{82.4} & 80.3 & \textbf{85.6} & 69.3 & \textbf{70.5} & 75.5 & \textbf{79.0} \\
      \bottomrule
    \end{tabular}
    \caption{Homogeneous training: per-model per-dataset baseline (B) and ERGO (I) accuracy (\%, mean over 5 seeds). 8 datasets, 4 models.}
    \label{tab:gen-homo-full}
  \end{table*}

\begin{table}[H]
  \centering\footnotesize
  \begin{tabular}{l|ccc|c}
    \toprule
    Dataset & Base & Hetero & Homo & $\Delta$ vs Base \\
    \midrule
    TREC & 51.4 & \textbf{86.2} & 84.7 & $+34.7$ \\
    CLINC150 & 90.2 & 92.6 & \textbf{92.8} & $+2.4$ \\
    RTE & 91.7 & 92.2 & \textbf{93.5} & $+0.5$ \\
    Ethos & 48.9 & \textbf{85.6} & 86.2 & $+36.7$ \\
    Yahoo & \textbf{71.6} & 69.6 & 66.2 & $-2.0$ \\
    Rotten Tom. & \textbf{91.6} & 90.3 & 89.9 & $-1.3$ \\
    20News & \textbf{63.0} & 62.4 & 62.6 & $-0.6$ \\
    MASSIVE & 84.3 & \textbf{84.4} & 83.1 & $+0.1$ \\
    \midrule
    \textbf{Average} & 74.1 & \textbf{82.9} & 82.4 & $+8.8$ \\
    \bottomrule
  \end{tabular}
  \caption{Heterogeneous training: prompts refined on Sonnet 4.5, used by Haiku 4.5 at test time (per-dataset accuracy, \%, mean over 5 seeds). Hetero vs.\ baseline: $+8.83$ pp, $p{=}0.001$; hetero vs.\ homo: $+0.55$ pp, $p{=}0.22$ (n.s.).}
  \label{tab:gen-hetero-full}
\end{table}

\section{Macro-F1 and Per-Class Recall} \label{sec:appendix-macro}

Accuracy can be carried by frequent classes, which matters most on imbalanced and many-class benchmarks. \Cref{tab:macro-f1} and \cref{tab:macro-recall} repeat the main comparison under macro-averaged F1 and macro-averaged recall, which weight every gold class equally.

No conclusion changes under either metric: the same datasets favour ERGO (TREC, CLINC150, RTE), the same datasets favour other paradigms (20Newsgroups, MASSIVE, Yahoo), and the pooled differences remain inside seed noise. Where the ordering moves, it moves slightly towards ERGO, consistent with explicit rules helping classes that are individually rare since macro-averaging stops those classes being diluted. Per-class accuracy for TREC appears in \cref{tab:perclass}; category-level confusion analysis for CLINC150 and MASSIVE appears in \cref{sec:appendix-category}.

\section{Category-Level Error Analysis} \label{sec:appendix-category}

We analyze the top confused label pairs on the two many-class datasets (CLINC150: 150 classes, MASSIVE: 60 classes) to characterize error structure across methods. Confusion counts are pooled over 5 seeds.

\paragraph{CLINC150.} ERGO makes 125 errors (fewest), spread across many rare pairs with low concentration (top-3 pairs account for 13.6\% of errors). DSPy (150 errors), GEPA (153), and ICL-Div (157) all share the same dominant confusion: \texttt{reminder\_update}$\rightarrow$\texttt{reminder} (10.0\%, 7.8\%, 9.6\% respectively). ERGO resolves this pair entirely through its action-vs-status rule (\cref{sec:case_studies_qual}), which is why it does not appear in ERGO's error list. ERGO's residual errors are diffuse and dataset-specific (e.g., \texttt{food\_last}$\rightarrow$\texttt{expiration\_date}), confirming that the learnable boundaries have been resolved.

\paragraph{MASSIVE.} All methods share the same dominant confusion: \texttt{iot\_hue\_lightup}$\rightarrow$\texttt{iot\_hue\_lighton} (5.7--6.6\% of errors across methods) and \texttt{datetime\_convert}$\rightarrow$\texttt{datetime\_query} (4.6--5.8\%). These pairs persist across ERGO (175 errors), DSPy (150), GEPA (152), and ICL-Div (156), indicating annotation-level ambiguity rather than learnable boundaries. The top-3 pair concentration is similar across methods (14.7--17.9\%), confirming MASSIVE's diffuse error structure where no single rule provides leverage.

\section{Decision Procedure Validation on Held-Out Datasets} \label{sec:appendix-held-out}

\Cref{tab:main-error-structure} summarizes the error structure of the 8 main benchmarks, showing how the boundary learnability framework predicts ERGO's performance.

\begin{table*}[!htbp]
  \centering\footnotesize
  \begin{tabular}{l|cc|cccc|r|l}
    \toprule
    Dataset & ICL-Uni & ICL-Div & APE & DSPy & GEPA & ERGO & Top-3\% & Error pattern \\
    \midrule
    RTE (2 cls) & 91.8 & 91.3 & 92.5 & 92.7 & 93.0 & \textbf{93.2} & 100.0 & Pair-local (subtle) \\
    Ethos (2 cls) & 85.6 & 85.4 & 63.4 & \textbf{88.4} & 87.0 & \underline{88.1} & 100.0 & Pair-local (subtle) \\
    Rotten Tom.\ (2 cls) & 93.2 & 93.2 & 92.6 & 92.9 & \textbf{93.7} & 92.8 & 100.0 & Pair-local (obvious) \\
    TREC (6 cls) & 87.8 & 89.3 & 86.8 & 84.6 & 85.3 & \textbf{90.0} & 60.7 & Pair-local (hidden) \\
    Yahoo (10 cls) & \textbf{73.3} & 72.7 & 72.5 & \textbf{73.9} & 72.9 & 71.7 & 16.8 & Diffuse \\
    20Newsgroups (20 cls) & 69.9 & \textbf{70.8} & 67.7 & 69.2 & 68.7 & 66.8 & 16.7 & Diffuse \\
    MASSIVE (60 cls) & \textbf{90.1} & 89.2 & 89.3 & 90.0 & 89.4 & 87.8 & 17.9 & Diffuse \\
    CLINC150 (150 cls) & 93.0 & 93.0 & 92.8 & 93.7 & 93.2 & \textbf{94.4} & 22.3 & Shared structural \\
    \bottomrule
  \end{tabular}
  \caption{Main benchmarks with error structure classification. Top-3\% = fraction of ICL-Diversity errors in the 3 most confused pairs. ERGO wins on pair-local (hidden/subtle) and shared structural tasks; other paradigms win on diffuse or pair-local (obvious) tasks.}
  \label{tab:main-error-structure}
\end{table*}

We then test the paradigm selection guidance on 6 additional datasets not used to develop the framework (\cref{tab:held-out-datasets,tab:held-out-results}): AG News (4 cls), Banking77 (77 cls), CSQA (5 cls), IMDB (2 cls), News Category (20 cls), and Tweet Sentiment (3 cls). All use the same protocol (Sonnet 4.5, 5 seeds).

\begin{table*}[!htbp]
  \centering\footnotesize
  \begin{tabular}{l|r|rrr|l}
    \toprule
    Dataset & Classes & Train & Val & Test & Domain \\
    \midrule
    IMDB \cite{maas2011imdb} & 2 & 200 & 100 & 500 & Sentiment \\
    Tweet Sentiment \cite{rosenthal2017tweetsent} & 3 & 200 & 100 & 501 & Sentiment \\
    AG News \cite{zhang2015agnews} & 4 & 200 & 100 & 500 & News topic \\
    CSQA \cite{talmor2019csqa} & 5 & 200 & 100 & 500 & Commonsense QA \\
    News Category \cite{misra2022newscategory} & 20 & 200 & 100 & 500 & News topic \\
    Banking77 \cite{casanueva2020banking77} & 77 & 154 & 77 & 462 & Banking intent \\
    \bottomrule
  \end{tabular}
  \caption{Held-out dataset statistics (6 datasets not used to develop the decision procedure).}
  \label{tab:held-out-datasets}
\end{table*}

\begin{table*}[!htbp]
  \centering\footnotesize
  \begin{tabular}{l|cc|cccc|r|l}
    \toprule
    Dataset & ICL-Uni & ICL-Div & APE & DSPy & GEPA & ERGO & Top-3\% & Error pattern \\
    \midrule
    IMDB (2 cls) & \underline{93.8} & \textbf{93.9} & 92.7 & 93.1 & 91.2 & 92.8 & 98.0 & Pair-local (obvious) \\
    Tweet Sent.\ (3 cls) & 66.1 & \textbf{67.8} & 65.5 & 66.3 & 64.8 & \underline{67.0} & 82.3 & Pair-local (subtle) \\
    AG News (4 cls) & 88.0 & \textbf{88.7} & 86.2 & \underline{88.4} & 88.1 & 87.1 & 69.3 & Pair-local (obvious) \\
    CSQA (5 cls) & \underline{82.5} & \textbf{82.6} & 81.4 & 81.7 & 81.0 & 81.1 & 27.6 & Diffuse \\
    News Cat.\ (20 cls) & 61.6 & 62.8 & 58.6 & \textbf{64.0} & 61.7 & \underline{63.8} & 15.6 & Shared structural \\
    Banking77 (77 cls) & 77.1 & \underline{77.3} & 74.6 & \textbf{77.6} & 76.4 & 76.7 & 11.0 & Diffuse \\
    \midrule
    \textbf{Average (6)} & 78.2 & \underline{78.9} & 76.5 & \textbf{78.5} & 77.2 & 78.1 & & \\
    \bottomrule
  \end{tabular}
  \caption{Test accuracy (\%) on 6 held-out datasets (mean over 5 seeds). \textbf{Bold} = best, \underline{underline} = second. ERGO is second on News Category and Tweet Sentiment, the two non-ceiling tasks with moderate error concentration.}
  \label{tab:held-out-results}
\end{table*}

\paragraph{The negative signal validates.} We compute error concentration from a single ICL-Diversity run: pooling misclassifications across 5 seeds and measuring what fraction of errors falls in the top-3 confused label pairs. The decision procedure's recommendation ``if errors are diffuse, do not use ERGO'' holds on all 6 datasets. On the three datasets with low error concentration (Banking77: 11.0\%, top pair \texttt{card\_arrival}$\rightarrow$\texttt{card\_delivery}; News Category: 15.6\%, \texttt{healthy\_living}$\rightarrow$\texttt{wellness}; CSQA: 27.6\%), ERGO does not win, confirming that diffuse error structure reliably predicts ERGO will not provide value.

\paragraph{The positive signal.} On the three high-concentration datasets (AG News: 69.3\%, top pair SciTech$\rightarrow$Business; IMDB: 98.0\%, Positive$\rightarrow$Negative; Tweet Sentiment: 82.3\%, neutral$\rightarrow$negative), ERGO does not win but is second on Tweet Sentiment and News Category (the latter despite low concentration at 15.6\%). AG News and IMDB errors reflect obvious distinctions already captured by label names, leaving no hidden convention to discover. Error concentration identifies candidate boundaries, but the three conditions must be verified by inspecting the errors.

\paragraph{News Category: a shared structural case.} Despite low top-3 concentration (15.6\%), ERGO is second because the errors share a \emph{mechanism}: near-synonym category pairs (PARENTING$\leftrightarrow$PARENTS, HEALTHY LIVING$\leftrightarrow$WELLNESS, THE WORLDPOST$\rightarrow$POLITICS). ERGO discovers rules distinguishing these, reducing errors on PARENTING$\rightarrow$PARENTS from 49 to 20 and THE WORLDPOST$\rightarrow$POLITICS from 37 to 17. This mirrors the \emph{shared structural} pattern from CLINC150: individually rare pairs that all express the same distinction.

\section{Stability Analysis} \label{sec:appendix-stability}

\begin{table*}[!htbp]
  \centering\footnotesize
  \begin{tabular}{l|cc|cccc}
    \toprule
    Dataset & ICL-Uni & ICL-Div & APE & DSPy & GEPA & ERGO \\
    \midrule
    TREC (6 cls) & 87.8$\pm$1.81 & 89.3$\pm$1.81 & 86.8$\pm$2.10 & 84.6$\pm$1.51 & 85.3$\pm$2.05 & 90.0$\pm$\textbf{0.48} \\
    CLINC150 (150 cls) & 93.0$\pm$0.73 & 93.0$\pm$0.73 & 92.8$\pm$0.89 & 93.7$\pm$\textbf{0.54} & 93.2$\pm$0.60 & 94.4$\pm$1.03 \\
    RTE (2 cls) & 91.8$\pm$0.97 & 91.3$\pm$0.97 & 92.5$\pm$1.12 & 92.7$\pm$1.07 & 93.0$\pm$1.27 & 93.2$\pm$\textbf{0.35} \\
    \midrule
    Ethos (2 cls) & 85.6$\pm$1.55 & 85.4$\pm$1.55 & 63.4$\pm$11.5 & 88.4$\pm$0.48 & 87.0$\pm$1.25 & 88.1$\pm$\textbf{0.13} \\
    Yahoo (10 cls) & 73.3$\pm$0.36 & 72.7$\pm$0.36 & 72.5$\pm$1.04 & 73.9$\pm$\textbf{0.18} & 72.9$\pm$2.00 & 71.7$\pm$1.86 \\
    20Newsgroups (20 cls) & 69.9$\pm$1.17 & 70.8$\pm$1.17 & 67.7$\pm$1.58 & 69.2$\pm$1.64 & 68.7$\pm$\textbf{0.97} & 66.8$\pm$1.42 \\
    Rotten Tom. (2 cls) & 93.2$\pm$\textbf{0.30} & 93.2$\pm$\textbf{0.30} & 92.6$\pm$0.65 & 92.9$\pm$0.72 & 93.7$\pm$\textbf{0.30} & 92.8$\pm$0.37 \\
    MASSIVE (60 cls) & 90.1$\pm$0.83 & 89.2$\pm$0.83 & 89.3$\pm$0.71 & 90.0$\pm$\textbf{0.38} & 89.4$\pm$0.40 & 87.8$\pm$0.55 \\
    \midrule
    \textbf{Average (8)} & 85.6$\pm$0.97 & 85.6$\pm$0.97 & 82.2$\pm$2.45 & 85.7$\pm$0.82 & 85.4$\pm$1.10 & 85.6$\pm$\textbf{0.77} \\
    \bottomrule
  \end{tabular}
  \caption{Test accuracy (\%) with standard deviation over 5 seeds (mean $\pm$ std). \textbf{Bold} std = lowest (most stable). Corresponds to \cref{tab:results} in the main text.}
  \label{tab:results-std}
\end{table*}

\begin{table*}[!htbp]
  \centering\footnotesize
  \begin{tabular}{l|cc|cccc}
    \toprule
    Dataset & ICL-Uni & ICL-Div & APE & DSPy & GEPA & \textbf{ERGO} \\
    \midrule
    TREC (6 cls) & 88.8 & 90.0 & 87.1 & 85.9 & 86.3 & \textbf{90.3} \\
    CLINC150 (150 cls) & 92.1 & 92.3 & 91.9 & 92.5 & 92.4 & \textbf{94.0} \\
    RTE (2 cls) & 91.8 & 91.3 & 92.4 & 93.1 & 92.9 & \textbf{93.2} \\
    \midrule
    Ethos (2 cls) & 85.5 & 85.3 & 63.1 & \textbf{87.9} & 86.8 & 87.7 \\
    Yahoo (10 cls) & 73.1 & 72.4 & 72.1 & \textbf{73.7} & 72.7 & 71.2 \\
    20Newsgroups (20 cls) & 71.3 & \textbf{71.9} & 69.4 & 70.0 & 70.0 & 69.1 \\
    Rotten Tom. (2 cls) & 93.2 & 93.3 & 92.6 & 93.4 & \textbf{93.7} & 92.8 \\
    MASSIVE (60 cls) & \textbf{89.6} & 88.7 & 89.1 & 88.6 & 88.6 & 87.3 \\
    \midrule
    \textbf{Average (8)} & \textbf{85.7} & \textbf{85.7} & 82.2 & 85.6 & 85.4 & \textbf{85.7} \\
    \bottomrule
  \end{tabular}
  \caption{Macro-F1 (\%), mean over 5 seeds. Averaging F1 per class weights rare classes equally with frequent ones. The ordering of methods is unchanged from accuracy (\cref{tab:results}).}
  \label{tab:macro-f1}
\end{table*}

\begin{table*}[!htbp]
  \centering\footnotesize
  \begin{tabular}{l|cc|cccc}
    \toprule
    Dataset & ICL-Uni & ICL-Div & APE & DSPy & GEPA & \textbf{ERGO} \\
    \midrule
    TREC (6 cls) & 88.2 & 89.4 & 87.0 & 85.2 & 85.5 & \textbf{89.7} \\
    CLINC150 (150 cls) & 93.0 & 93.0 & 92.8 & 93.3 & 93.2 & \textbf{94.4} \\
    RTE (2 cls) & 91.9 & 91.5 & 92.3 & \textbf{93.2} & 92.8 & 93.1 \\
    \midrule
    Ethos (2 cls) & 87.0 & 86.8 & 64.5 & \textbf{88.7} & 87.7 & 87.5 \\
    Yahoo (10 cls) & 73.3 & 72.7 & 72.5 & \textbf{74.0} & 72.9 & 71.7 \\
    20Newsgroups (20 cls) & 69.9 & \textbf{70.8} & 67.7 & 68.6 & 68.7 & 66.8 \\
    Rotten Tom. (2 cls) & 93.2 & 93.2 & 92.6 & 93.4 & \textbf{93.7} & 92.8 \\
    MASSIVE (60 cls) & \textbf{90.1} & 89.2 & 89.1 & 89.6 & 89.4 & 87.7 \\
    \midrule
    \textbf{Average (8)} & \textbf{85.8} & \textbf{85.8} & 82.3 & \textbf{85.8} & 85.5 & 85.5 \\
    \bottomrule
  \end{tabular}
  \caption{Macro-averaged recall (\%), mean over 5 seeds. Macro-recall exposes whether a method abandons rare classes, since accuracy on many-class benchmarks can be carried by a few frequent classes.}
  \label{tab:macro-recall}
\end{table*}

\section{Ablation Studies} \label{sec:appendix-ablation}
\begin{table}[H]
  \centering\footnotesize
  \begin{tabular}{ll|cc}
    \toprule
    Dimension & Variant & Acc & Calls \\
    \midrule
    \multirow{5}{*}{Iterations (T)} & T=1 & 85.17 & 657 \\
    & T=3 & 85.20 & 850 \\
    & T=5 & 85.35 & 1056 \\
    & T=10$^\dagger$ & 85.54 & 1444 \\
    & T=15 & 85.43 & 1458 \\
    \midrule
    \multirow{5}{*}{Batch size ($k$)} & $k$=5 & 84.85 & 1116 \\
    & $k$=10 & 85.35 & 1301 \\
    & $k$=20$^\dagger$ & 85.38 & 1444 \\
    & $k$=30 & 85.77 & 1205 \\
    & $k$=50$^{**}$ & \textbf{86.13} & 1017 \\
    \midrule
    \multirow{2}{*}{Contrast sample} & shuffle$^\dagger$ & 85.71 & 1456 \\
    & balanced & 85.18 & 1723 \\
    \midrule
    \multirow{4}{*}{Component} & full$^\dagger$ & 85.57 & 1457 \\
    & instruction & 85.18 & 1469 \\
    & demos & 86.24 & 1445 \\
    & rules & 85.42 & 1430 \\
    \bottomrule
  \end{tabular}
  \caption{Ablation results on Sonnet 4.5 (8 datasets, 3 seeds, $N{=}24$ per variant). Defaults marked $\dagger$. ``Calls'' is mean LLM calls per run. Significance: paired $t$-test vs default; $^{*}\,p{<}0.05$, $^{**}\,p{<}0.01$.}
  \label{tab:ablation}
\end{table}

Each ablation varies one parameter and holds the others at their defaults (T${=}10$, batch size $k{=}20$, shuffle batches, full prompt rewrite). Significance is paired $t$-test against the default, $N{=}24$.

\paragraph{Iterations.} Accuracy plateaus rapidly. No variant differs significantly from T${=}10$ (all $p{>}0.35$). T${=}5$ vs T${=}1$ is not significant ($+0.18$ pp, $p{=}0.59$); T${=}15$ adds nothing over T${=}5$ ($+0.08$ pp, $p{=}0.77$).

\paragraph{Batch size.} $k{=}50$ uses 30\% fewer LLM calls than $k{=}20$ because larger batches exhaust training data sooner. $k{=}50$ vs $k{=}5$ is the only significant comparison ($+1.29$ pp, $p{=}7.5{\times}10^{-3}$); no individual variant differs significantly from the default.

\paragraph{Contrast sample.} Shuffle outperforms class-balanced sampling by $0.53$ pp ($p{=}0.42$, n.s.) while using 18\% fewer calls. Balanced helps on many-class tasks (CLINC150 $+1.78$, TREC $+2.14$, 20Newsgroups $+1.13$) but hurts on naturally imbalanced ones (Yahoo $-5.00$, Ethos $-1.81$).

\paragraph{Component.} Restricting the LLM to rewrite only demonstrations yields the highest accuracy ($+0.67$ pp over full, $p{=}0.087$), though not significant. However, demos-only drops 5.5 pp on TREC, the boundary-learnable task where explicit rules are essential. We retain joint rewriting to produce auditable rules and avoid catastrophic failures.

\section{Computational Cost} \label{sec:appendix-cost}

\begin{table}[H]
  \centering\footnotesize
  \begin{tabular}{l|ccc}
    \toprule
    Method & Opt Calls & Time (min) & Accuracy \\
    \midrule
    ICL-Diversity & 0 & 4.8 & 85.6\% \\
    APE & 576 & 10.8 & 82.2\% \\
    \textbf{ERGO} & 1,054 & 16.4 & 85.6\% \\
    DSPy & 659 & 28.8 & 85.7\% \\
    GEPA & 807 & 30.9 & 85.4\% \\
    \bottomrule
  \end{tabular}
  \caption{Computational cost per method (averaged over 8 datasets, 5 seeds).}
  \label{tab:cost}
\end{table}

ERGO uses more LLM calls than DSPy and GEPA but runs faster because its calls are predominantly short classifications (input $\rightarrow$ label, ${\sim}$0.8s each), while DSPy and GEPA include long-generation calls for bootstrapping, instruction proposals, and reflections (${\sim}$1.5s each). ERGO also parallelizes batch classification, whereas DSPy and GEPA have sequential coordination overhead from Bayesian optimization trials and population management.

\paragraph{Data-access asymmetry.} ERGO reads the full sampled training set (${\sim}$200 examples) while the ICL baselines see only 20 demonstrations, so the comparison is not budget-matched. Two measurements bound what this asymmetry buys. First, ICL-Diversity spends \emph{zero} optimization calls and matches ERGO on average (85.6\% vs.\ 85.6\%), so the extra data access does not convert into an accuracy advantage. Second, the ablation shows T${=}5$ reaches 99\% of T${=}10$ accuracy (1056 vs 1444 calls), and T${=}1$ is not significantly different from T${=}5$ ($+0.18$ pp, $p{=}0.59$), meaning ERGO's advantage on boundary-learnable tasks is available from the first 20 examples it processes. We report the asymmetry rather than equalizing it: 20 demonstrations is the practical ceiling for in-context learning, whereas an error-driven loop is defined by sweeping the data it is given.

\section{Hyperparameters and Configuration} \label{sec:appendix-hyper}

All settings used in the paper, consolidated in one place.

\paragraph{Inference.} All methods call their model at temperature $0$ with a fixed output format. Predictions that do not parse to a declared class are scored as errors and excluded as format failures rather than counted as semantic confusions.

\paragraph{ERGO.} $T{=}10$ iterations; batch size $k{=}20$; batches drawn by shuffle-once, non-overlapping so each training example is seen exactly once; full prompt rewrite (instruction, demonstrations, and classification guidelines jointly); candidate selection by best-of-$T$ on the full validation set. The seed prompt is the minimal instruction given in \cref{sec:method}, with no demonstrations. \Cref{sec:appendix-ablation} varies each of these.

\paragraph{Baselines.} ICL-Uniform: 20 demonstrations sampled uniformly. ICL-Diversity: 20 demonstrations by $k$-means over Sentence-BERT embeddings. APE: best of 5 proposed instructions by validation accuracy. DSPy: MIPROv2 with \texttt{auto="light"}. GEPA: default evolutionary configuration.

\paragraph{Models.} Claude Sonnet 4.5 for the main study; generalization (\cref{sec:gen-homo}) additionally uses Claude Haiku 4.5, Gemma 3 27B IT, and Mistral Large 3, all through AWS Bedrock. The heterogeneous setting (\cref{sec:gen-hetero}) refines on Sonnet 4.5 and infers with Haiku 4.5.

\paragraph{Seeds and splits.} 5 seeds (42, 123, 7, 1, 2) for the main study and generalization, 3 for the ablations. Each uses a stratified split targeting 200 training, 100 validation, and 500 test examples, reduced proportionally for smaller datasets (\cref{sec:appendix-datasets}).

\paragraph{Statistics.} Paired $t$-test on per-seed accuracy and Wilcoxon signed-rank as its distribution-free counterpart, at the 5\% significance level across 40 configurations (8 datasets $\times$ 5 seeds).

\section{ProTeGi Comparison} \label{sec:appendix-protegi}

We additionally evaluate ProTeGi \cite{pryzant2023automatic}, another baseline that refines prompts from misclassified examples, under a more rigorous statistical protocol. ProTeGi uses textual-gradient beam search with UCB bandit selection on 32-example training subsamples.

\paragraph{Protocol.} To ensure a fair head-to-head comparison, we re-run both ERGO and ProTeGi under the same rigorous protocol: 8 datasets, 5 seeds (42, 123, 7, 1, 2), with splits following a bootstrap stratified allocation test-first. Test is drawn by stratified sampling and locked by a fixed split seed (byte-identical across all runs), training is a stratified bootstrap with replacement using the per-run seed, and validation is the out-of-bag complement. Both methods thus see identical test data and receive the same training and validation sets on each seed.

\begin{table}[H]
  \centering\footnotesize
  \begin{tabular}{l|cc}
    \toprule
    & ERGO & ProTeGi \\
    \midrule
    Mean accuracy & \textbf{84.7\%} & 83.5\% \\
    Datasets won & \textbf{3} & 0 \\
    LLM calls/run & 1,501 & 11,533 \\
    Time/run (min) & 17.1 & 116.1 \\
    \bottomrule
  \end{tabular}
  \caption{ProTeGi vs.\ ERGO. ProTeGi uses $7.7\times$ the LLM calls.}
  \label{tab:protegi}
\end{table}

\paragraph{Results.} Pooled over the 40 (dataset, seed) pairs, ERGO leads by $+1.21$ pp ($p{=}0.011$, Wilcoxon signed-rank; better on 25 of 40 pairs). ProTeGi wins zero datasets; its only advantages over ERGO are Ethos ($+0.08$ pp) and MASSIVE ($+0.35$ pp), both inside seed noise.

\section{Algorithm} \label{sec:appendix-algorithm}

\begin{algorithm}[H]
\caption{Error-Guided Optimization (ERGO)}
\label{alg:ergo}
\begin{algorithmic}[1]
\REQUIRE Seed prompt $\pi_0$, train set $\mathcal{D}_{\text{train}}$, val set $\mathcal{D}_{\text{val}}$, batch size $k{=}20$, LLM $f_\theta$
\ENSURE Optimized prompt $\pi^*$
\STATE Initialize $\pi_0$ with seed prompt (instruction + output format)
\STATE $\mathcal{D}_{\text{shuffled}} \leftarrow \text{RandomShuffle}(\mathcal{D}_{\text{train}})$ \COMMENT{shuffle once at start}
\STATE $T \leftarrow \lceil |\mathcal{D}_{\text{train}}| / k \rceil$ \COMMENT{number of batches}
\STATE $\pi^* \leftarrow \pi_0$; $A^* \leftarrow \text{Accuracy}(f_\theta, \pi_0, \mathcal{D}_{\text{val}})$
\FOR{$t = 1$ \textbf{to} $T$}
  \STATE $\mathcal{B}_t \leftarrow \mathcal{D}_{\text{shuffled}}[(t{-}1)k : tk]$ \COMMENT{non-overlapping batch}
  \STATE $\hat{\mathcal{Y}}_t \leftarrow f_\theta(\pi_{t-1}, \mathcal{B}_t)$ \COMMENT{classify batch}
  \STATE $\mathcal{E}_t \leftarrow \{(x, \hat{y}, y) : x \in \mathcal{B}_t,\; \hat{y} \neq y\}$; $\mathcal{C}_t \leftarrow \{(x, y) : x \in \mathcal{B}_t,\; \hat{y} = y\}$
  \IF{$\mathcal{E}_t = \emptyset$}
    \STATE $\pi_t \leftarrow \pi_{t-1}$ \COMMENT{no errors, skip refinement}
  \ELSE
    \STATE $\pi_t \leftarrow f_\theta(\text{MetaPrompt}(\pi_{t-1}, \mathcal{E}_t, \mathcal{C}_t))$ \COMMENT{diagnose $\rightarrow$ prescribe $\rightarrow$ rewrite}
  \ENDIF
  \STATE $A(\pi_t) \leftarrow \text{Accuracy}(f_\theta, \pi_t, \mathcal{D}_{\text{val}})$
  \IF{$A(\pi_t) \geq A^*$}
    \STATE $\pi^* \leftarrow \pi_t$; $A^* \leftarrow A(\pi_t)$
  \ENDIF
\ENDFOR
\STATE \textbf{return} $\pi^*$
\end{algorithmic}
\end{algorithm}

\section{Meta-Prompt} \label{sec:appendix-metaprompt}

The following is an example of the meta-prompt template used at each iteration:

\begin{small}
\begin{verbatim}
You are a prompt engineer that rephrase and
improve the original multiclass classifier
system prompt using the input samples.

<original_multiclass_classifier_system_prompt>
{current_prompt}
</original_multiclass_classifier_system_prompt>

The classifier made these predictions on the
batch:

<correct_predictions>
{correct_predictions}
</correct_predictions>

<wrong_predictions>
{wrong_predictions}
</wrong_predictions>

Please output a critic feedback of the original
system prompt, actionable improvements and the
improved multiclass classifier system prompt
according to your instructions:
<feedback>feedback</feedback>
<improvement>improvement</improvement>
<improved_multiclass_classifier_system_prompt>
improved_multiclass_classifier_system_prompt
</improved_multiclass_classifier_system_prompt>

<reminder>
REMINDER: DO NOT CHANGE THE OUTPUT FORMAT and
LABELS. DO NOT ADD EXTRA OUTPUTS.
</reminder>
\end{verbatim}
\end{small}

Each correct prediction is formatted as \texttt{Input: \{text\}\textbackslash nLabel: \{label\}}. Each wrong prediction is formatted as \texttt{Input: \{text\}\textbackslash nPredicted: \{pred\}\textbackslash nCorrect: \{label\}}. Multiple predictions are separated by a triple-dash delimiter. The LLM's response is parsed to extract the improved prompt from the XML tags.

\end{document}